# Energy Efficiency in Robotics Software: A Systematic Literature Review (2020-2024)

Aryan Gupta
arygupta@umich.edu
ORCID: 0009-0001-8179-5145

## Abstract

This study presents a systematic literature review of software-level approaches to energy efficiency in robotics published from 2020 through 2024, updating and extending pre-2020 evidence. An automated-but-audited pipeline combined Google Scholar seeding, backward/forward snowballing, and large-language-model (LLM) assistance for screening and data extraction, with ~10% human audits at each automated step and consensus-with-tie-breaks for full-text decisions. The final corpus comprises 79 peer-reviewed studies analyzed across application domain, metrics, evaluation type, energy models, major energy consumers, software technique families, and energy–quality trade-offs. Industrial settings dominate (31.6%), followed by exploration (25.3%). Motors/actuators are identified as the primary consumer in 68.4% of studies, with computing/controllers a distant second (13.9%). Simulation-only evaluations remain most common (51.9%), though hybrid evaluations are frequent (25.3%). Representational (physics-grounded) energy models predominate (87.3%). Motion and trajectory optimization is the leading technique family (69.6%), often paired with learning/prediction (40.5%) and computation allocation/scheduling (26.6%); power management/idle control (11.4%) and communication/data efficiency (3.8%) are comparatively underexplored. Reporting is heterogeneous: composite objectives that include energy are most common, while task-normalized and performance-per-energy metrics appear less often, limiting cross-paper comparability. The review offers a minimal reporting checklist (e.g., total energy and average power plus a task-normalized metric and clear baselines) and highlights opportunities in cross-layer designs and in quantifying non-performance trade-offs (accuracy, stability). A replication package with code, prompts, and frozen datasets accompanies the review.

## 1. Introduction

In recent years, the robotics community has developed a growing awareness of the energy implications of software. As robotic systems become increasingly autonomous and embedded in real-world settings; spanning logistics, manufacturing, search and rescue, and personal assistance; concerns about energy consumption have intensified. While advances in hardware enable more energy-efficient actuators, sensors, and computing platforms, the software that governs robotic behavior remains a significant and under-exploited lever for improvement [1,2].

This intersection between software design and energy efficiency is particularly salient in mobile and embedded robotics, where limited battery capacity and real-time constraints put a premium on intelligent resource management. Software determines not only when and how hardware is activated, but also how navigation, perception, scheduling, communication, and control decisions are made, each with a measurable impact on energy consumption.

In 2020, Swanborn and Malavolta [2] conducted one of the first systematic literature reviews (SLRs) to examine energy efficiency in robotics from a software perspective. Their work synthesized insights from 17 primary studies and proposed a classification framework spanning energy-saving techniques, energy models, evaluation strategies, and trade-offs with other qualities. While foundational, their review necessarily reflected the state of the field at that time.

Since 2020, building on gaps identified by Swanborn and Malavolta, the field has evolved. This study observed increased availability of energy-aware middleware, simulation platforms with energy-profiling capabilities, and design-time tools for estimating the energy impact of software decisions. Concurrently, emerging application domains such as swarm robotics, human-robot interaction, and adaptive mission planning have introduced new constraints and optimization challenges [3,4].

This study responds to that need. This study presents a systematic literature review of studies published between 2020 and 2024 that address energy efficiency in robotics software. Building on the methodology developed by Swanborn and Malavolta, this study modernized and automated the search-and-screening pipeline (LLM-assisted) with manual audits, and applied both backward and forward snowballing to capture a comprehensive, current body of work. Ultimately, this study identified 79 peer-reviewed primary studies, which this study analyzed across application domain, used metrics, evaluation type, energy models, major energy consumers, software-level energy-saving techniques, and trade-offs with other system qualities. Methodological details (databases, search strings, screening procedures, automation and auditing) are provided in Section 2. A replication package with code, prompts, and frozen datasets (see Section 2) was also released.

The contributions are threefold:
C1 - Updated synthesis. An up-to-date synthesis of the field based on a substantially larger corpus (2020-2024), reflecting the evolving landscape of robotics and energy-aware software.
C2 - Reproducible methodology. A reproducible, extensible methodology that leverages LLM-assisted screening and extraction with auditing mechanisms, accompanied by an open replication package.
C3 - Categorized insights. A structured set of findings on how energy efficiency is addressed in robotic software today, highlighting dominant approaches, gaps, and promising directions for future work.

The primary audience for this review includes both researchers seeking to extend energy-efficient robotics methodologies and practitioners looking for software strategies to reduce energy consumption in real deployments.

This study is structured as follows: Section 2 details the methodology, encompassing search, screening, and data extraction. Section 3 presents the primary findings from the 79 selected studies. Section 4 explores trends, identifies gaps, discusses implications, and notes limitations. Finally, Section 5 concludes this study.

# 2. Methodology

This systematic literature review (SLR) extends Swanborn and Malavolta's 2020 study to literature published from January 1, 2020 (papers in the year 2020 that appeared in Swanborn and Malavolta's study were excluded manually at the start) through December 31, 2024. This study combines established SLR protocols with substantial automation to improve scalability, reproducibility, and transparency, following accepted guidelines for secondary studies in software engineering and robotics [5,6]. A complete replication package, including code, selection results, extraction scripts, and frozen raw/filtered datasets, is available at https://github.com/guptaaryanr/slr_robotics_replication_package.

## 2.1 Research Goal and Question

The goal is to identify, analyze, and classify contemporary research on energy efficiency in robotics software since 2020, with emphasis on software-level strategies.

*[RQ1] What is the current state of the art in analyzing and improving the energy efficiency of robotics software (2020-2024)?*

Operationalization of RQ1 via data items appears in Section 2.3.

## 2.2 Study Design and Workflow

This study follows established SLR guidance [5,6], building on Swanborn and Malavolta [2] and enhancing replicability via automation while maintaining human oversight. Large language models (LLMs), reference APIs, and full-text parsing tools are integrated across search, filtering, screening, and extraction [7,8]. Decision quality was ensured by conducting human audits of ~10% of the papers chosen at every step of LLM decision making and instituting manual tie-breaks for papers for which the decision was not constant across multiple instances of the LLM consensus method.

**Figure 1: Workflow Chart**

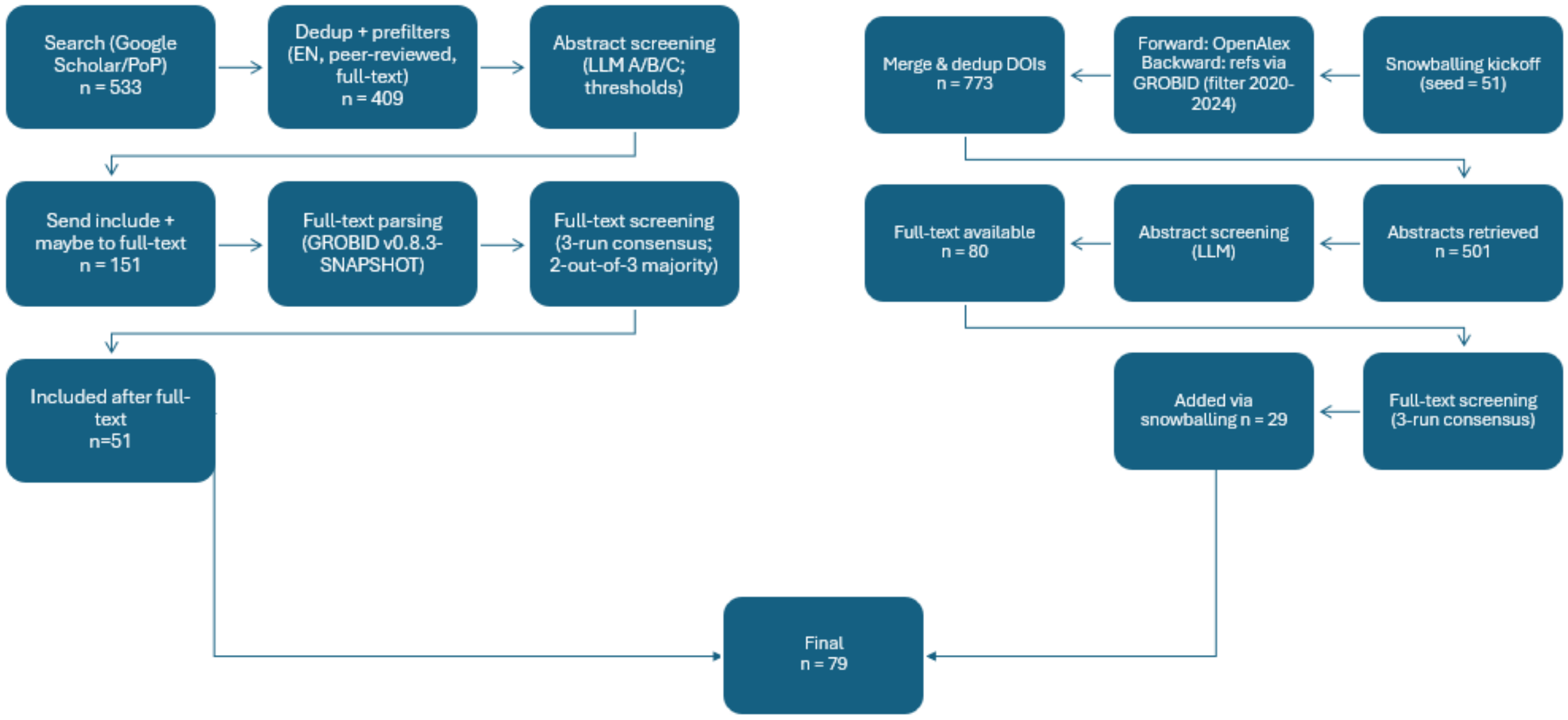

### 2.2.1 Initial Search and Filtering

An initial corpus using Publish or Perish on Google Scholar [9] with the title-restricted query was constructed:

*intitle:robot AND (intitle:power OR intitle:green OR intitle:energy OR intitle:battery) AND software*

This yielded 533 records. This study prioritized precision via title restriction and relied on snowballing (Section 2.2.4) to improve recall. Google Scholar was chosen for its breadth and accessibility, mitigating single-source risks via deduplication and snowballing [5,6].

Syntactic and semantic duplicates, non-English papers, non-peer-reviewed content, secondary/tertiary studies, non-archival formats, and papers without accessible full text were removed. After filtering, 409 papers remained and were downloaded for automated analysis.

### 2.2.2 Filtering via Inclusion and Exclusion Criteria

This study adapted Swanborn and Malavolta's criteria [2] for a larger, more recent corpus.

Inclusion criteria:
(i1) strictly robotics application;
(i2) explicit energy efficiency as a central focus;

(i3) software as the primary lever of implementation for energy efficiency (≥50% of contribution at software level);
(i4) empirical evaluation/validation/application;
(i5) peer-reviewed;
(i6) English.

Exclusion criteria:
(e1) energy efficiency not tied to software improvements;
(e2) energy only an illustrative example;
(e3) secondary/tertiary studies;
(e4) non-archival formats;
(e5) no full text available for download.

Borderline cases (e.g., workshop papers, journal-extended versions, and preprints) were resolved via precedence rules. Journal versions were preferred if available for a paper with conference or workshop papers. Preprints were summarily rejected.

**2.2.3 Abstract and Full-Text Screening**

Abstracts of the 409 records were screened using GPT-4 (gpt-4 with temperature=0) to label include/exclude/maybe with confidence scores, following emerging work on LLM-assisted screening [7,8]. A random audit subset of ~10% of the papers (40 papers) was manually reviewed to estimate accuracy and reduce false negatives. Papers labeled as 'maybe' were retained for automated full text analysis after a stricter manual review of all of them. Papers that failed to run due to extraction or filepath issues were manually analyzed. This yielded 151 records labeled include or maybe for full-text screening.

Full texts were parsed with GROBID [10] and screened via a three-pass consensus protocol: three independent LLM runs using GPT-4 (gpt-4-1106-preview for its larger context window with temperature=0), with decisions made by majority vote and low-confidence ties or decision conflicts escalated to manual review [11,12]. All final inclusion decisions received human verification. 51 papers passed full-text screening.

**2.2.4 Snowballing**

Backward snowballing parsed references of the 51 studies with GROBID [10,13]; items were filtered by year (2020-2024) and deduplicated (DOI, normalized title). Forward snowballing used OpenAlex to retrieve citing works through the cited_by_reference_url endpoint [13,14].

From both arms, 773 DOIs were identified, but only 501 abstracts were retrievable via OpenAlex (others lacked metadata or were inaccessible). Abstract screening retained 88 candidates, of which 80 had accessible full text, and full-text screening retained 28 studies. The same abstract and full text screening procedure as listed above (Section 2.2.3) was applied. The final corpus totals 79 primary studies.

**Table 1: Stage-wise counts, cross-referenced with Figure 1**

| Stage | Count |
|---|---|
| Initial search | 533 |
| Duplicate removal and manual review | 409 |
| Automated abstract screening and manual audit | 151 |
| Automated full text screening with consensus and manual audit | 51 |
| Snowballing (forward/backward) | 773 |
| Snowballed paper retrieval | 501 |
| Snowballed abstract screening with manual audit | 80 |
| Snowballed full text screening | 28 |
| **Final included studies** | **51 + 28 = 79** |

## 2.3 Data Extraction

Full texts were manually skimmed for validity (some early-access papers that were not found initially were also removed in this stage to maintain the narrow scope of this study) and then parsed with GROBID (v0.8.3-SNAPSHOT) [10]. From each TEI XML, we concatenated text from sections whose headers matched Methods/Approach, Results/Experiments, and Discussion/Threats; if these were not detected, we fell back to the first ~10k tokens of the body text.

For each paper, we executed three independent LLM extractions with prompt variants A/B/C at fixed parameters (gpt-4-1106-preview, temperature = 0, seed = 42) and enforced a function-calling JSON schema (OpenAI "tools" with forced tool_choice) so that outputs adhered to a closed set of labels per field. The system prompt instructed exact enum usage and an "other if uncertain" default; user prompts A/B/C varied wording and examples to reduce prompt-locking effects while keeping label semantics identical.

Field schema and category induction (in-line synthesis): Early free-text runs produced large numbers of unique strings per field, so we switched to in-line synthesis: we first explored uniques and then crystallized closed label sets directly in the extraction schema, iteratively refining them until coverage stabilized. The exploration script reports per-column uniqueness to guide this induction. The final extractor emits seven items (adapted from [2]) with these label sets:

- Used Metrics (category of metric, not raw unit): aggregate energy/power; task-normalized energy; performance per energy; relative change vs baseline; composite objective incl. energy; physics-based integrals; validation statistics; operational proxies.
- Application Domains: robot exploration; industrial; service/domestic; aerial; aquatic; modular; additive manufacturing; IoT power; swarm/multi-robot; mixed.
- Major Energy Consumers: motors/actuators; sensors; computing/controllers; communication subsystem; battery/power electronics; mechanical motion pattern; idle/stand-by overhead.
- Evaluation Type: simulation; physical; hybrid.
- Energy Model: abstract; representational (terminology aligned to Swanborn and Malavolta's model typology). [2]
- Software Techniques (multi-label): power management & idle control; motion & trajectory optimization; computation allocation & scheduling; learning/predictive optimization; communication & data efficiency; hardware/morphology & harvesting; (optionally) other/hybrid.
- QA Trade-offs: e.g., performance vs energy; reliability/accuracy vs energy; coverage/mission quality vs energy; safety/stability vs energy; maintainability/code complexity vs energy; cost/hardware vs energy (enumerated in schema).

The schema also records confidence and an optional quote per field evidence map for auditability.

Consensus and auditing: We applied a field-level consensus across the three runs. For single-label fields, a 2-of-3 majority was accepted; if no majority emerged, we selected the confidence-weighted winner (sum of per-run confidences). When neither condition was met, we assigned "other/none" (when available in the schema), flagged the record, and routed it to manual review. For the list-valued techniques field, we kept any item appearing in ≥2 of 3 runs. The metric (being category-like but occasionally noisy) was taken from the highest-confidence run. We stored vote tallies, a review flag, and the max confidence for each record.

All per-run JSONs, the consensus JSON, and the aggregated CSV are preserved for replication; prompts A/B/C are versioned with the code.

Units and label normalization policy: Because the metric field is categorical (e.g., "aggregate energy/power"), we preserve the paper-reported units in text and normalize only when unambiguous: power → W, energy → Wh (SI prefixes normalized, e.g., kWh); W/h (rate-of-change of power) was not used. Where metrics were reported solely in Joules (J) or other units, we preserved the reported units and added a normalized value only when a clear conversion basis existed. Minor label variants/typos (e.g., "maintainabilility") were corrected during schema induction; originals are kept alongside normalized labels in the replication package.

Rationale and safeguards: We adopted this closed-schema, multi-run, consensus-plus-audit design to balance LLM efficiency with reliability. Prior studies on LLM extraction highlight both promise and risks (format drift, hallucinations); enforcing a schema, varying prompts, and majority/confidence voting are recommended mitigations in emerging practice [15,16]. Human oversight remains essential; our flagged cases and evidence quotes support targeted checks rather than exhaustive re-reads.

## 2.4 Final Corpus

The final corpus comprises 79 primary studies (2020-2024), identified through direct search and snowballing.

**Table 2: Primary Studies**

| Title | Year | Author(s) | Publication |
|---|---|---|---|
| Energy-Aware Path Planning for Autonomous Mobile Robot Navigation | 2020 | Renan Maidana, Roger Granada, Darlan Jurak, Mauricio Magnaguagno, Felipe Meneguzzi, Alexandre Amory | Conference |
| An experimental energy consumption comparison between trajectories generated by using the cart-table model and an optimization approach for the Bioloid robot | 2020 | Tacué, Jeison; Rengifo, Carlos; Bravo, Diego | Journal |
| Energy Comparison of Controllers Used for a Differential Drive Wheeled Mobile Robot | 2020 | Stefek, Alexandr; Pham, Thuan Van; Krivanek, Vaclav; Pham, Khac Lam | Journal |
| Energy-Conscientious Trajectory Planning for an Autonomous Mobile Robot in an Asymmetric Task Space | 2020 | Bakshi, Soovadeep; Feng, Tianheng; Yan, Zeyu; Ma, Zheren; Chen, Dongmei | Journal |
| Least-Energy Path Planning With Building Accurate Power Consumption Model of Rotary Unmanned Aerial Vehicle | 2020 | Hong, Dooyoung; Lee, Seonhoon; Cho, Young Hoo; Baek, Donkyu; Kim, Jaemin; Chang, Naehyuck | Journal |
| Mathematical methods for planning energy-efficient motion path of the manipulator anthropomorphic robot for the typical obstacles | 2020 | Petrenko, V I; Tebueva, F B; Antonov, V O; Gurchinsky, M M | Journal |
| Minimizing the Energy Consumption for a Hexapod Robot Based on Optimal Force Distribution | 2020 | Wang, Guanyu; Ding, Liang; Gao, Haibo; Deng, Zongquan; Liu, Zhen; Yu, Haitao | Journal |
| Multi-Robot Energy-Efficient Coverage Control | 2020 | TURANLI, Mert; TEMELTAS, Hakan | Journal |
| with Hopfield Networks | | | |
| Reinforcement Learning-Based Energy-Aware Area Coverage for Reconfigurable hRombo Tiling Robot | 2020 | Le, Anh Vu; Parween, Rizuwana; Kyaw, Phone Thiha; Mohan, Rajesh Elara; Minh, Tran Hoang Quang; Borusu, Charan Satya Chandra Sairam | Journal |
| Adaptive Floor Cleaning Strategy by Human Density Surveillance Mapping with a Reconfigurable Multi-Purpose Service Robot | 2021 | Sivanantham, Vinu; Le, Anh Vu; Shi, Yuyao; Elara, Mohan Rajesh; Sheu, Bing J. | Journal |
| An efficient transmission algorithm for power grid data suitable for autonomous multi-robot systems | 2021 | Chen, Xiaoyan; Liang, Wei; Zhou, Xinlian; Jiang, Dingchao; Kui, Xiaoyan; Li, Kuang-Ching | Journal |
| An Energy-Efficient Communication Scheme for Multi-robot Coordination Deployed for Search and Rescue Operations | 2021 | Rajesh, M.; Nagaraja, S. R. | Book Chapter |
| Applied energy optimization of multi-robot systems through motion parameter tuning | 2021 | Hovgard, Mattias; Lennartson, Bengt; Bengtsson, Kristofer | Journal |
| Battery Charge Dispatching in Multi-robot Systems | 2021 | Xing, Zichao; Wu, Weimin; Niu, Haoyi; Hu, Ruifen | Conference |
| Coverage Path Planning Using Reinforcement Learning-Based TSP for hTetran—A Polyabolo-Inspired Self-Reconfigurable Tiling Robot | 2021 | Le, Anh Vu; Veerajagadheswar, Prabakaran; Thiha Kyaw, Phone; Elara, Mohan Rajesh; Nhan, Nguyen Huu Khanh | Journal |
| Energy Consumption of Control Schemes for the Pioneer 3DX Mobile Robot: Models and Evaluation | 2021 | Jaiem, Lotfi; Crestani, Didier; Lapierre, Lionel; Druon, Sébastien | Journal |
| Energy-Efficient Mobile Robot Control via Run-time Monitoring of Environmental Complexity and Computing Workload | 2021 | Mohamed, Sherif A.S.; Haghbayan, Mohammad-Hashem; Miele, Antonio; Mutlu, Onur; Plosila, Juha | Conference |
| Finite-Horizon Kinetic Energy Optimization of a Redundant Space Manipulator | 2021 | Tringali, Alessandro; Cocuzza, Silvio | Journal |
| Optimal Trajectory Planning for Wheeled Mobile Robots under Localization Uncertainty and Energy Efficiency Constraints | 2021 | Zhang, Xiaolong; Huang, Yu; Rong, Youmin; Li, Gen; Wang, Hui; Liu, Chao | Journal |
| Optimization of energy consumption for hexapod robot following inclined path using nontraditional gait | 2021 | Beaber, Sameh; Khadr, Mohamed Sh.; AbdelHamid, Ahmed Y.; Abou Elyazed, Maged M. | Conference |
| Optimization of Fuzzy Logic Controller Used for a Differential Drive Wheeled Mobile Robot | 2021 | Štefek, Alexandr; Pham, Van Thuan; Krivanek, Vaclav; Pham, Khac Lam | Journal |
| Parallel Swarm Intelligent Motion Planning with Energy-Balanced for Multirobot in Obstacle Environment | 2021 | Su, Shoubao; Zhao, Wei; Wang, Chishe | Journal |
| Performance Analysis of Task Allocation for Mobile Robot Exploration Under Energy Constraints | 2021 | Soni, Ankit | Conference |
| PID++: A Computationally Lightweight Humanoid Motion Control Algorithm | 2021 | Arciuolo, Thomas F.; Faezipour, Miad | Journal |
| Power Line Inspection Tasks With Multi-Aerial Robot Systems Via Signal Temporal Logic | 2021 | Silano, Giuseppe; Baca, Tomas; Penicka, Robert; Liuzza, Davide; Saska, Martin | Journal |
| Specifications | | | |
| Predicting the Energy Consumption of a Robot in an Exploration Task Using Optimized Neural Networks | 2021 | Caballero, Liesle; Perafan, Álvaro; Rinaldy, Martha; Percybrooks, Winston | Journal |
| Reinforcement Learning-Based Complete Area Coverage Path Planning for a Modified hTrihex Robot | 2021 | Apuroop, Koppaka Ganesh Sai; Le, Anh Vu; Elara, Mohan Rajesh; Sheu, Bing J. | Journal |
| Resource-Constrained Scheduling for Multi-Robot Cooperative Three-Dimensional Printing | 2021 | Poudel, Laxmi; Zhou, Wenchao; Sha, Zhenghui | Journal |
| A novel energy consumption model for autonomous mobile robot | 2022 | GÜRGÖZE, GÜRKAN; TÜRKOĞLU, İBRAHİM | Journal |
| A Novel Resolution Scheme of Time-Energy Optimal Trajectory for Precise Acceleration Controlled Industrial Robot Using Neural Networks | 2022 | Hou, Renluan; Niu, Jianwei; Guo, Yuliang; Ren, Tao; Yu, Xiaolong; Han, Bing; Ma, Qun | Journal |
| A Novel, Energy-Efficient Smart Speed Adaptation Based on the Gini Coefficient in Autonomous Mobile Robots | 2022 | Gürgöze, Gürkan; Türkoğlu, İbrahim | Journal |
| An energy-efficient and self-triggered control method for robot swarm networking systems | 2022 | Byun, Heejung; Yang, Soo-mi | Journal |
| Decentralized and Centralized Planning for Multi-Robot Additive Manufacturing | 2022 | Poudel, Laxmi; Elagandula, Saivipulteja; Zhou, Wenchao; Sha, Zhenghui | Journal |
| Energy Saving Planner Model via Differential Evolutionary Algorithm for Bionic Palletizing Robot | 2022 | Deng, Yi; Zhou, Tao; Zhao, Guojin; Zhu, Kuihu; Xu, Zhaixin; Liu, Hai | Journal |
| Energy-Efficient Path Planning of Reconfigurable Robots in Complex Environments | 2022 | Kyaw, Phone Thiha; Le, Anh Vu; Veerajagadheswar, Prabakaran; Elara, Mohan Rajesh; Thu, Theint Theint; Nhan, Nguyen Huu Khanh; Van Duc, Phan; Vu, Minh Bui | Journal |
| Evaluation of Selected Algorithms for Air Pollution Source Localisation Using Drones | 2022 | Suchanek, Grzegorz; Wołoszyn, Jerzy; Gołaś, Andrzej | Journal |
| Intelligent Energy Management System for Mobile Robot | 2022 | Lee, Min-Fan Ricky; Nugroho, Asep | Journal |
| Minimalist Coverage and Energy-Aware Tour Planning for a Mobile Robot | 2022 | Ghosh, Anirban; Dutta, Ayan; Sotolongo, Brian | Conference |
| Modeling and Simulation Analysis of Energy Self-sustainment Behavior Decision in Robot Ecosphere | 2022 | Nenghao, Hu; Taibo, Li; Hongwei, Liu; Lei, Xu | Conference |
| On the Effect of Heterogeneous Robot Fleets on Smart Warehouses' Order Time, Energy, and Operating Costs | 2022 | Oliveira, George S.; Carvalho, Jonata T.; Plentz, Patricia D. M. | Conference |
| Optimal Energy Consumption Path Planning for Quadrotor UAV Transmission Tower Inspection Based on Simulated Annealing Algorithm | 2022 | Wu, Min; Chen, Wuhua; Tian, Xiaohong | Journal |
| Optimal Path Planning for Wireless Power Transfer Robot Using Area Division Deep | 2022 | Xing, Yuan; Young, Riley; Nguyen, Giaolong; Lefebvre, Maxwell; Zhao, | Journal |
| Reinforcement Learning | | Tianchi; Pan, Haowen; Dong, Liang | |
| Performance optimization of fuel cell hybrid power robot based on power demand prediction and model evaluation | 2022 | Lü, Xueqin; Deng, Ruiyu; Chen, Chao; Wu, Yinbo; Meng, Ruidong; Long, Liyuan | Journal |
| Simulation to Real: Learning Energy-Efficient Slithering Gaits for a Snake-Like Robot | 2022 | Bing, Zhenshan; Cheng, Long; Huang, Kai; Knoll, Alois | Journal |
| TDE2-MBRL: Energy-exchange Dynamics Learning with Task Decomposition for Spring-loaded Bipedal Robot Locomotion | 2022 | Kuo, Cheng-Yu; Shin, Hirofumi; Kamioka, Takumi; Matsubara, Takamitsu | Conference |
| Towards High-Quality Battery Life for Autonomous Mobile Robot Fleets | 2022 | Chavan, Akshar Shravan; Brocanelli, Marco | Conference |
| WearROBOT: An Energy Conservative Wearable Obstacle Detection Robot With LP Multi-Commodity Graph | 2022 | Okafor, Kennedy Chinedu; Longe, Omowunmi Mary | Journal |
| A metaheuristic approach to optimal morphology in reconfigurable tiling robots | 2023 | Kalimuthu, Manivannan; Pathmakumar, Thejus; Hayat, Abdullah Aamir; Elara, Mohan Rajesh; Wood, Kristin Lee | Journal |
| An Efficient Framework for Autonomous UAV Missions in Partially-Unknown GNSS-Denied Environments | 2023 | Mugnai, Michael; Teppati Losé, Massimo; Herrera-Alarcón, Edwin; Baris, Gabriele; Satler, Massimo; Avizzano, Carlo | Journal |
| Automated method based on a neural network model for searching energy-efficient complex movement trajectories of industrial robot in a differentiated technological process | 2023 | Gorkavyy, Mikhail A.; Gorkavyy, Aleksandr I.; Egorova, Valeria P.; Melnichenko, Markel A. | Journal |
| Collision-Free Navigation in Human-Following Task Using a Cognitive Robotic System on Differential Drive Vehicles | 2023 | Dang, Chien Van; Ahn, Heungju; Kim, Jong-Wook; Lee, Sang C. | Journal |
| Energy-efficient and quality-aware part placement in robotic additive manufacturing | 2023 | Ghungrad, Suyog; Mohammed, Abdullah; Haghighi, Azadeh | Journal |
| Energy-Efficient Blockchain-Enabled Multi-Robot Coordination for Information Gathering: Theory and Experiments | 2023 | Castellon, Cesar E.; Khatib, Tamim; Roy, Swapnoneel; Dutta, Ayan; Kreidl, O. Patrick; Bölöni, Ladislau | Journal |
| Energy-saving control of rolling speed for spherical robot based on regenerative damping | 2023 | Li, Yansheng; Yang, Meimei; Wei, Bo; Zhang, Yi | Journal |
| Investigation on the Mobile Wheeled Robot in Terms of Energy Consumption, Travelling Time and Path Matching Accuracy | 2023 | Szeląg, Piotr; Dudzik, Sebastian; Podsiedlik, Anna | Journal |
| Optimization of Energy Consumption of Industrial Robots Using Classical PID and MPC Controllers | 2023 | Benotsmane, Rabab; Kovács, György | Journal |
| Optimize Path Planning for Drone-based Wireless Power Transfer System by Categorized Reinforcement Learning | 2023 | Xing, Yuan; Verma, Abhishek | Conference |
| Optimizing Robotic Task Sequencing and Trajectory Planning on the Basis of Deep Reinforcement Learning | 2023 | Dong, Xiaoting; Wan, Guangxi; Zeng, Peng; Song, Chunhe; Cui, Shijie | Journal |
| Overcoming Obstacles With a Reconfigurable | 2023 | Simhon, Or; Karni, Zohar; Berman, Sigal; | Journal |
| Robot Using Deep Reinforcement Learning Based on a Mechanical Work-Energy Reward Function | | Zarrouk, David | |
| Power Consumption Modeling of Wheeled Mobile Robots With Multiple Driving Modes | 2023 | Kumar, Pushpendra; Bensekrane, Ismail; Merzouki, Rochdi | Journal |
| The Energy Efficiency Multi-Robot System and Disinfection Service Robot Development in Large-Scale Complex Environment | 2023 | Chen, Chin-Sheng; Lin, Feng-Chieh; Lin, Chia-Jen | Journal |
| The Need for Task-Specific Execution in Robot Manipulation: Skill Design for Energy-Efficient Control | 2023 | Deroo, Boris; Pousett, Brendan; Aertbeliën, Erwin; Decré, Wilm; Bruyninckx, Herman | Conference |
| A Data-Driven Method for Predicting and Optimizing Industrial Robot Energy Consumption Under Unknown Load Conditions | 2024 | Chang, Qing; Yuan, Tiantian; Li, Haifeng; Chen, Yuxiang; Wang, Xuehao; Gao, Sen; Ren, Hongsheng; Zhao, Xiangyun; Wang, Lingyu | Journal |
| A Maintenance-Aware Approach for Sustainable Autonomous Mobile Robot Fleet Management | 2024 | Atik, Syeda Tanjila; Chavan, Akshar Shravan; Grosu, Daniel; Brocanelli, Marco | Journal |
| BN-LSTM-based energy consumption modeling approach for an industrial robot manipulator | 2024 | Lin, Hsien-I; Mandal, Raja; Wibowo, Fauzy Satrio | Journal |
| Cognitive Model Predictive Learning Cooperative Control to Optimize Electric Power Consumption and User-Friendliness in Human–Robot Co-manipulation | 2024 | Rahman, S. M. Mizanoor | Book Chapter |
| Coordinated Ship Welding with Optimal Lazy Robot Ratio and Energy Consumption via Reinforcement Learning | 2024 | Yu, Rui; Chen, Yang-Yang | Journal |
| Design of a robust intelligent controller based neural network for trajectory tracking of high-speed wheeled robots | 2024 | Xue, Wenkui; Zhou, Baozhi; Chen, Fenghua; Ghaderpour, Ebrahim; Mohammadzadeh, Ardashir | Journal |
| Energy Consumption Minimization of Quadruped Robot Based on Reinforcement Learning of DDPG Algorithm | 2024 | Yan, Zhenzhuo; Ji, Hongwei; Chang, Qing | Journal |
| Energy efficient robot operations by adaptive control schemes | 2024 | Choi, Minje; Park, Seongjin; Lee, Ryujeong; Kim, Sion; Kwak, Juhyeon; Lee, Seungjae | Journal |
| Managing Energy Consumption of Linear Delta Robots Using Neural Network Models | 2024 | Vodovozov, Valery; Lehtla, Madis; Raud, Zoja; Semjonova, Natalia; Petlenkov, Eduard | Journal |
| Multiobjective Energy Consumption Optimization of a Flying–Walking Power Transmission Line Inspection Robot during Flight Missions Using Improved NSGA-II | 2024 | Wang, Yanqi; Qin, Xinyan; Jia, Wenxing; Lei, Jin; Wang, Dexin; Feng, Tianming; Zeng, Yujie; Song, Jie | Journal |
| Omnidirectional AGV Path Planning Based on Improved Genetic Algorithm | 2024 | Niu, Qinyu; Fu, Yao; Dong, Xinwei | Journal |
| Research the Potential for Energy Saving and Maximizing Productivity in Rectilinear Transitions of a Collaborative Robot | 2024 | Gorkavyy, Mikhail; Ivanov, Yuriy; Melnichenko, Markel | Conference |
| Robot Environment Modeling and Motion Control Approach for Sustainable Energy Savings in Mobile Robot Landmine Surveillance Mission | 2024 | Nagapriya, C. N.; Ashok, S. Denis | Book Chapter |
| Spike-based high energy efficiency and accuracy tracker for Robot | 2024 | Qu, Jinye; Gao, Zeyu; Li, Yi; Lu, Yanfeng; Qiao, Hong | Conference |
| The development of a neural network surrogate model for estimating energy and travel time for a collaborative robot | 2024 | Bouchra, Khoumeri; Abdelhadi, Abedou; Mehdi, Gaham; Afaf, Ammar; Isra, Rached; Taqwa, Mezaache | Conference |
| Three-dimensional spatial energy-quality map construction for optimal robot placement in multi-robot additive manufacturing | 2024 | Ghungrad, Suyog; Haghighi, Azadeh | Journal |
| Three-Tiered Controller for Obstacle Avoidance in a PV Panel-Powered Wheeled Mobile Robot: Considering Actuators and Power Electronics Stages | 2024 | Reyes-Reyes, Erik; Santiago-Nogales, Benjamin Natanael; Silva-Ortigoza, Ramón; Marciano-Melchor, Magdalena; García-Sánchez, José Rafael; Orta-Quintana, Ángel Adrián; Silva-Ortigoza, Gilberto; Taud, Hind; Hernández-Bolaños, Miguel | Journal |

# 3. Results

Insights gained from the analysis carried out on the extracted data from the primary studies listed are synthesized and described here.

## 3.1 Used Metrics

Across N = 79 post-2020 studies the energy-related outcomes were grouped into six core categories and two auxiliary (validation statistics and other) buckets. The most frequent form is a composite objective including energy, i.e., energy combined with other terms such as time, accuracy, or stability in a joint cost, which appears in 27 studies. Reporting of aggregate energy/power (episode energy in J/Wh or mean power in W, including current as a proxy) is also common (21 studies), as is relative change versus a baseline (e.g., % energy savings; 16 studies). Fewer papers provide task-normalized energy (e.g., J/m, CoT; 4 studies) or performance per energy (e.g., FPS/W; 4 studies). Physics-based integrals (e.g., kinetic-energy integrals along a trajectory) are rare (2 studies). Validation statistics (e.g., %RAE, r, IAE; 3 studies) were also tracked and an "other" category for idiosyncratic measures (2 studies); these are auxiliary and not treated as energy metrics in the headline distribution. Domains were coded as mutually exclusive; totals therefore sum to the number of included studies.

Definitions used for coding:

- Composite objective including energy: weighted or multi-objective formulations that mix energy with other terms (e.g., time-energy cost, energy-accuracy, stability-economy) [17,21,28,30,31,35,38,39,41,43,44,50,57,58,61,63,65-67,70,76,77,83,84,89,90,94].
- Aggregate energy/power: total energy over a run (J/Wh) or average power (W); current (A) counted as a proxy when voltage is known or stated [18,19,22,24,27,32,37,47,52,55,56,59,62,71,73,75,80,85,86,91,92].
- Relative change versus baseline: percent/ratio improvement in energy relative to a stated baseline [20,23,25,26,29,45,48,49,68,72,79,81,82,87,93,95].
- Task-normalized energy: energy per unit task, such as J/m, J/(kg·m), area/J; includes Cost of Transport (CoT) [33,36,46,51].

- Performance per energy: throughput or quality per unit energy/power (e.g., FPS/W) [40,53,60,88].
- Physics-based integrals: trajectory-level integrals (e.g., kinetic-energy integral) used as energy proxies [34,74].

Context vs. pre-2020: Swanborn & Malavolta highlighted FPS/W, J/m (incl. CoT), and aggregate energy/power as the most comparable forms before 2020 [2]; the post-2020 corpus shows a shift toward composite and baseline-relative reporting, with task-normalized and performance-per-energy metrics appearing less often. This has implications for comparability across 79 studies and tasks.

**Figure 2: Distribution of Metrics**

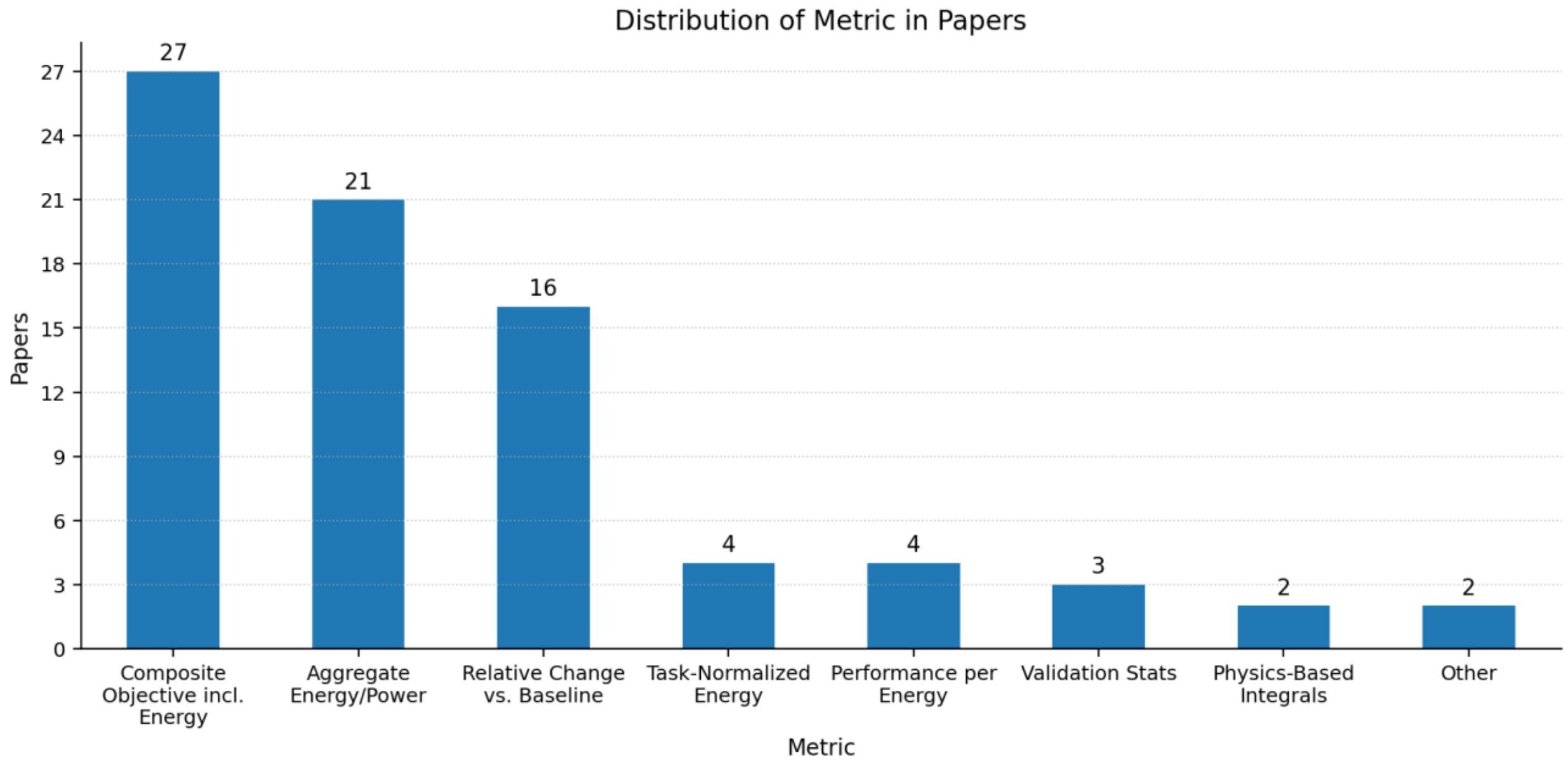

## 3.2 Application Domains

Across the corpus, industrial applications are most common (25/79, 31.6%), followed by robot exploration (20/79, 25.3%) and service/domestic settings (13/79, 16.5%). Less frequent are swarm or multi-robot (7/79, 8.9%), aerial (5/79, 6.3%), IoT power (4/79, 5.1%), additive manufacturing (4/79, 5.1%), and modular robotics (1/79, 1.3%). Domains were coded as mutually exclusive; totals therefore sum to the number of included studies [17-95].

Context vs. pre-2020: Compared to Swanborn & Malavolta (≤2020), who found robot exploration to be the most frequent domain with industrial a distant second, the post-2020 literature shifts toward industrial deployments, with exploration still prominent but no longer dominant [2]. This suggests growing attention to energy efficiency within production and manufacturing contexts, while exploration remains an active testbed where efficiency directly extends operational autonomy.

**Figure 3: Distribution of Application Domain**

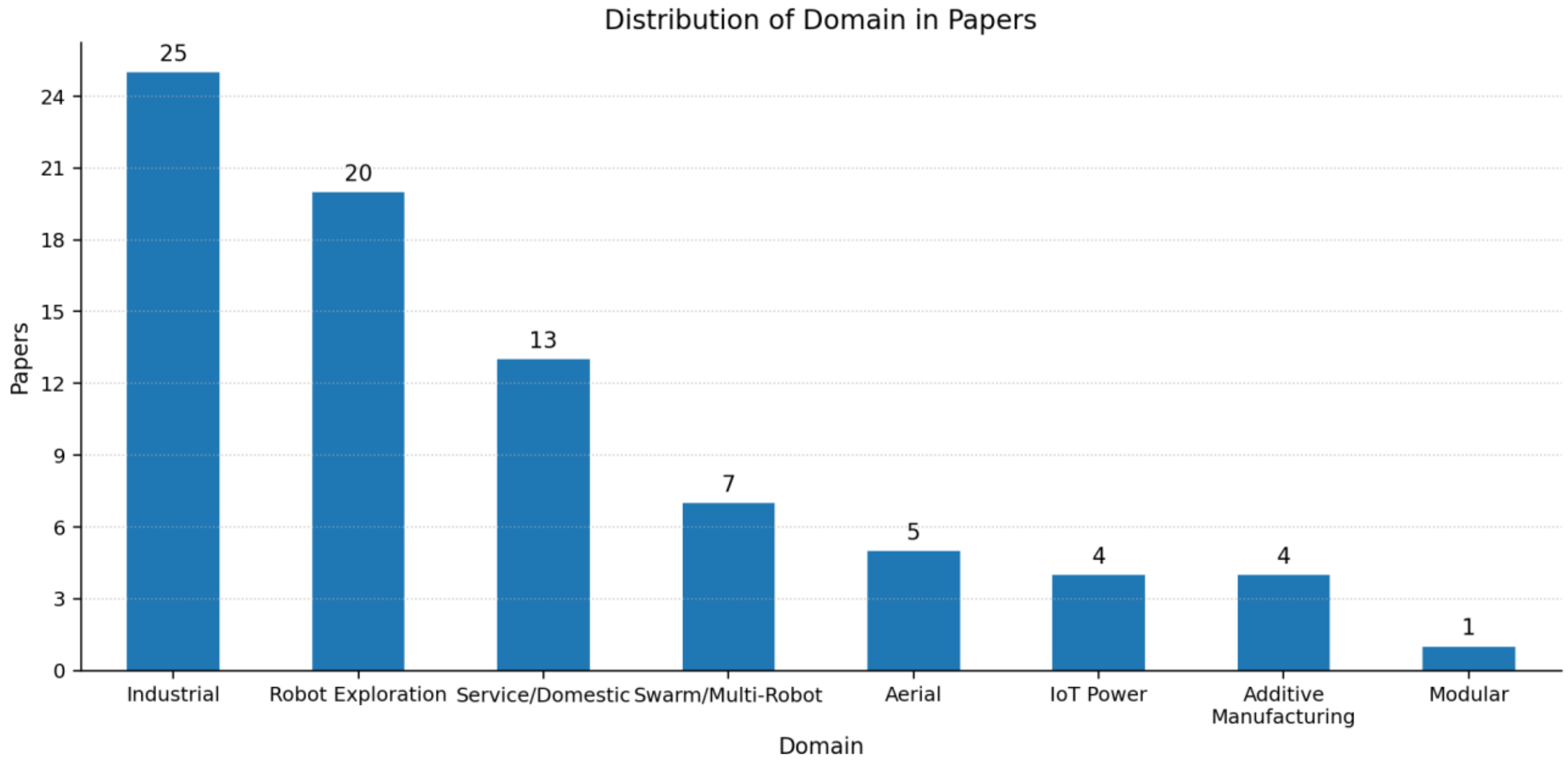

## 3.3 Identified Major Energy Consumers

Across the corpus, motors and actuators are most often identified as the primary energy sink (54/79, 68.4%), followed by computing and controllers (11/79, 13.9%). Far fewer studies single out sensors (5/79, 6.3%) or the communication subsystem (5/79, 6.3%). Only a small number cite the mechanical motion pattern, i.e., movement style such as frequent starts/stops or sharp turns, as the dominant consumer (2/79, 2.5%), or battery and power electronics (2/79, 2.5%). One primary consumer was coded per study; totals therefore equal the number of included studies.

Definitions used for coding:

- Motors and actuators: propulsion and manipulation actuation (including drivetrain and servos) [17,19-25,28,34-38,40-47,49-52,55,56,58-60,65,66,68,70-72,74,75,77-80,82,84,86-91,93-95].
- Computing and controllers: CPU/GPU/accelerators, embedded control units, and on-board scheduling/estimation loads [18,33,48,53,54,61,69,76,81,83,92].
- Sensors: perception stack hardware (e.g., cameras, LiDAR, IMU) and their acquisition pipelines [32,39,62,64,85].
- Communication subsystem: radios and network interfaces (Wi-Fi, cellular, mesh) and associated protocols [26,27,57,67,73].
- Mechanical motion pattern: energy attributable to the executed motion profile (e.g., stop-and-go, turning angle, acceleration/deceleration) [30,63].
- Battery and power electronics: DC-DC converters, drivers, and battery-management circuitry [29,31].

Context vs. pre-2020: Swanborn & Malavolta grouped "major consumers" by cause (inefficient procedure, movement, hardware) [2]. The hardware-subsystem view yields a clearer picture of where energy is actually spent, but it leads to a similar conclusion: actuation dominates, with computing a meaningful secondary sink. This is consistent with typical robot power budgets and explains why many software-level optimizations ultimately target actuator duty cycles, motion planning, or controller efficiency.

**Figure 4: Distribution of Identified Major Energy Consumers**

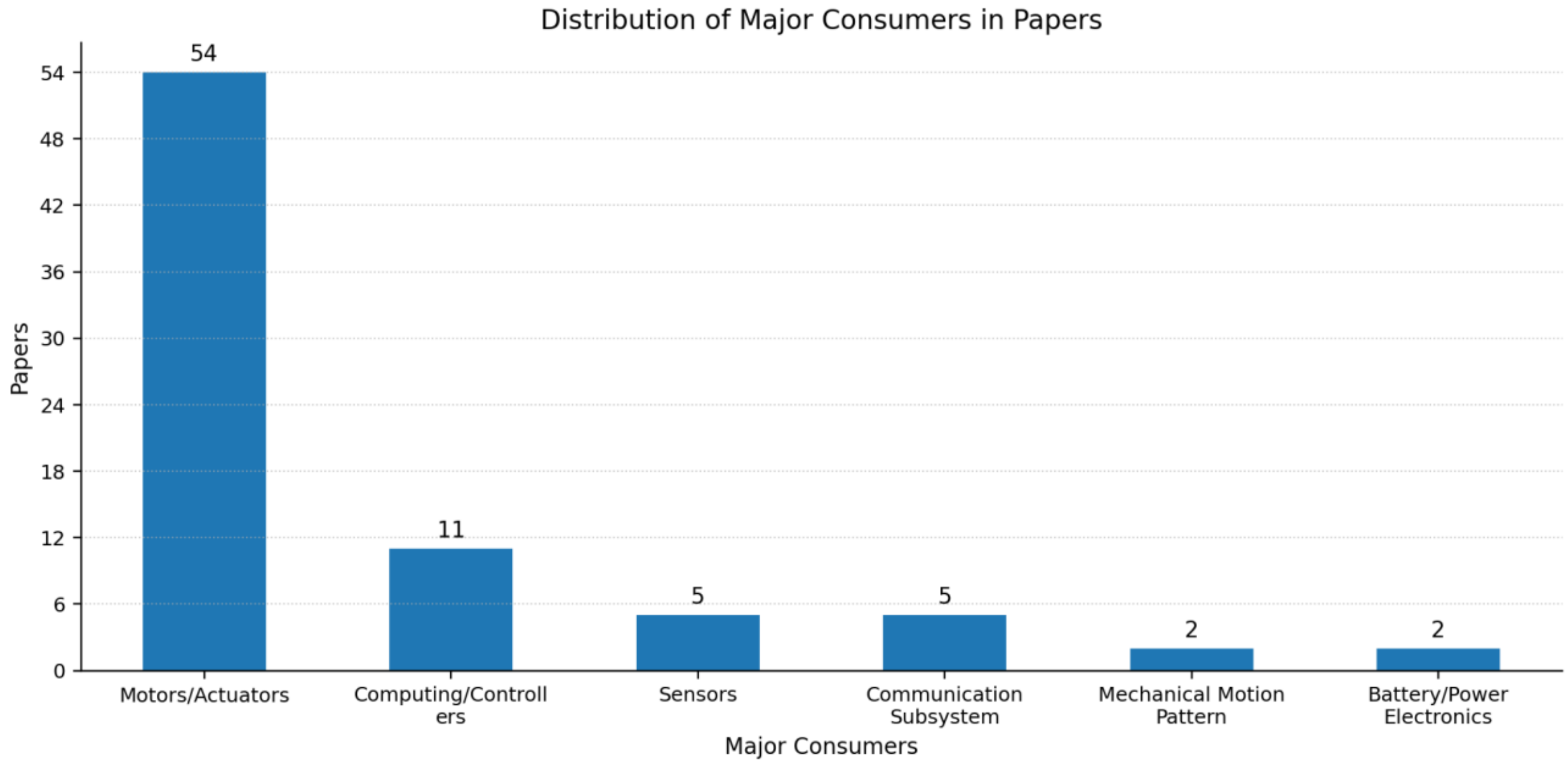

## 3.4 Evaluation Types

All included studies reported at least one experimental evaluation (79/79). Using mutually exclusive categories; simulation (only), physical (only), and hybrid (both); the distribution is: simulation (41/79, 51.9%), hybrid (20/79, 25.3%), and physical (18/79, 22.8%). Aggregating by modality, 77.2% of studies involved a simulated component (simulation + hybrid), while 48.1% included a physical component (physical + hybrid) [17-95].

Definitions used for coding:

- Simulation: experiments conducted entirely in a simulator.
- Physical: experiments conducted entirely on real robots/hardware.
- Hybrid: at least one simulated and one physical experiment reported for the same study.

Context vs. pre-2020: Swanborn & Malavolta reported that most studies relied on simulation [2], with a minority using physical or hybrid evaluations; one study had no experiment. In the 2020-2024 corpus, simulation remains the most common, but hybrid evaluations are more frequent and all studies include an experiment, suggesting stronger attention to empirical validation overall.

**Figure 5: Distribution of Evaluation Types**

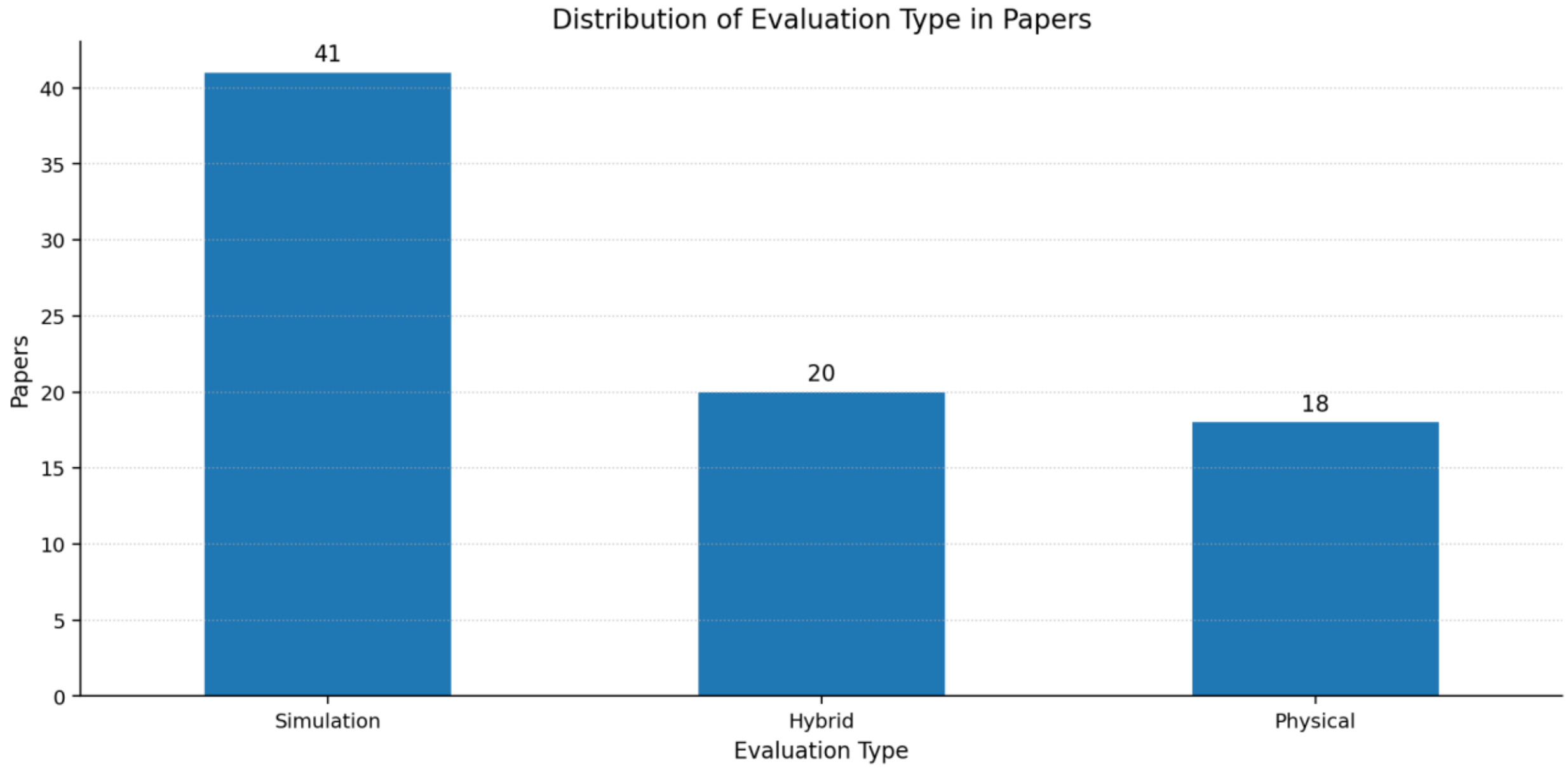

## 3.5 Energy Models

The energy model used (if any) was classified in each study as representational, abstract, or other. Most studies adopt representational models, physically grounded formulations intended to approximate real-world energy, typically parameterized by robot mass and geometry with friction, torque, acceleration/deceleration, drivetrain losses, and idle power terms (69/79, 87.3%). Abstract models, task-level heuristics that trade absolute fidelity for relative comparisons (e.g., grid costs where distance, turns, and stops incur fixed "energy" tokens), are rare (3/79, 3.8%). The remaining studies fall into other (7/79, 8.9%), covering cases with no explicit energy model or idiosyncratic/data-driven estimates that do not match the two main categories. One label was assigned per study.

Definitions used for coding:

- Representational: physics-based, parameterized energy models designed for comparability with real robots/sensors; often used inside simulators or for controller/trajectory evaluation [17-20,22,23,25-28,31-52,54-62,65,66,68-76,78-81,83-95].
- Abstract: simplified task-level models used to compare alternatives when only relative energy differences matter [21,63,77].
- Other: no model reported, or a black-box/empirical estimate that does not map cleanly to the above (e.g., black-box NN estimates without physical grounding) [24,29,30,53,64,67,82].

Context vs. pre-2020: Swanborn & Malavolta reported 11 representational, 4 none, and 2 abstract studies (one unassigned) [2]. In the post-2020 corpus the share of representational models is markedly higher, while abstract models remain uncommon; studies without a clear model are a small minority (the other bucket). This aligns with the broader shift toward simulation-heavy and hybrid evaluations noted in Section 3.4.

**Figure 6: Distribution of Energy Models**

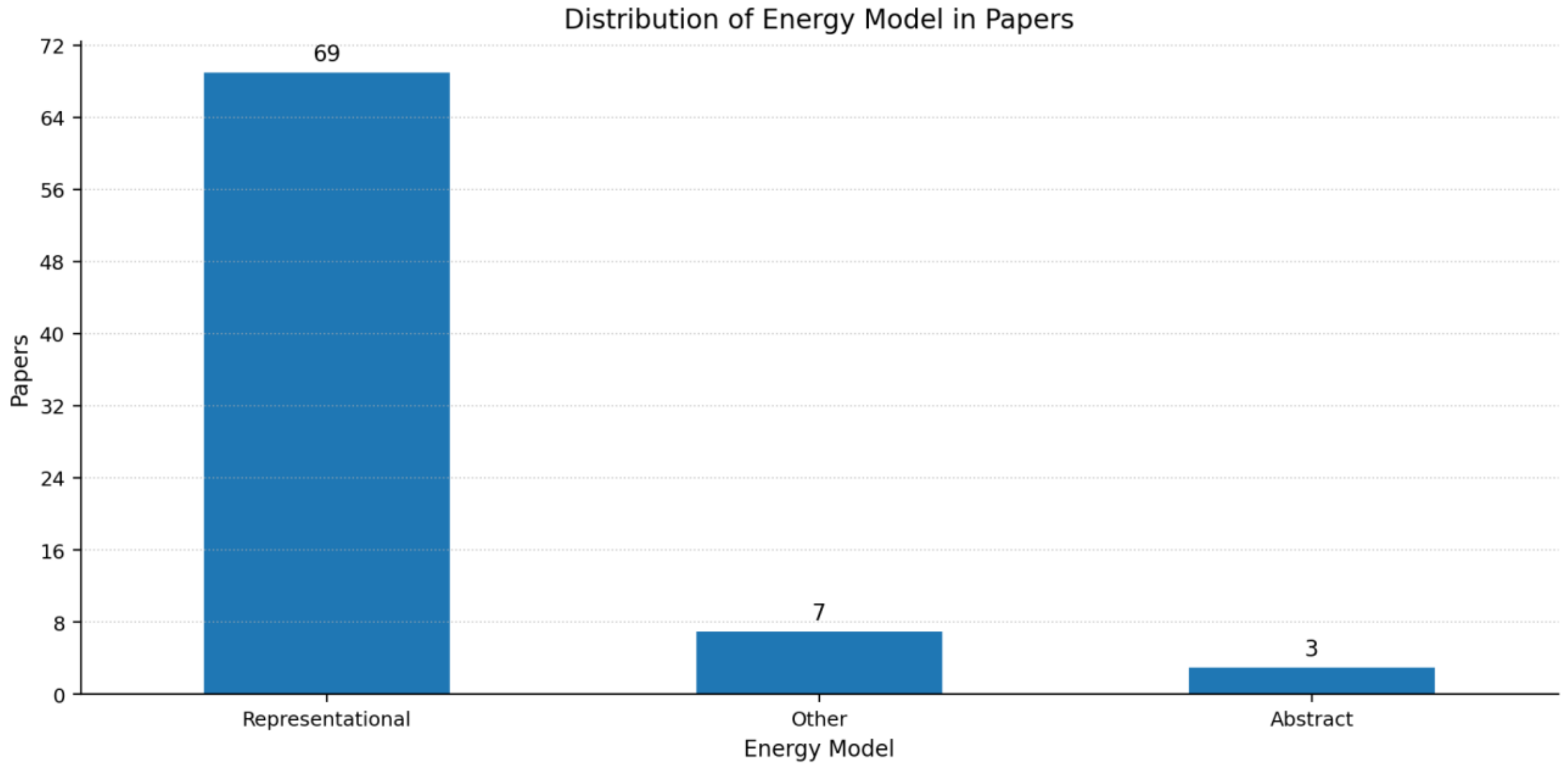

### 3.6 Techniques for Energy Efficiency

Each study was coded for the presence of up to six technique families (multi-label allowed). Across 79 studies, 121 family assignments were observed (mean 1.53 families per study), so totals below exceed 79 by design. A "hardware/morphology & harvesting" family was initially included; it had no instances in the corpus and is therefore omitted from the results.

(1) Motion & trajectory optimization - 55/79 (69.6%):
This family targets the dominant consumer, motors/actuators, by reducing actuation effort. Typical tactics include: (a) energy-aware global/local planning that shortens paths; (b) smoothing to reduce stop-start patterns and jerk; (c) turn minimization and smaller steering angles; (d) speed capping or cruise control to avoid costly accelerations; (e) obstacle-avoidance that anticipates rather than reacts; (f) multi-goal routing (e.g., TSP-style ordering) to minimize travel. Reported metrics are often task-normalized (e.g., J/m, CoT) or relative energy savings [17-25,28,30,31,34-41,43-47,49-51,53,55,56,58-60,64-66,68,70-72,75,77,80,82-84,86-91,94,95].

(2) Learning or predictive optimization - 32/79 (40.5%):
Studies use learned or model-based predictors to choose low-energy actions or reject tasks expected to exceed an energy budget. This study found (a) policy learning for energy-aware navigation or manipulation; (b) model-predictive control with an explicit energy term; (c) surrogate models that estimate energy from telemetry and guide planning; and (d) thresholding/admission control based on predicted energy cost. Effects are commonly reported as % energy reduction vs. a baseline or as improved J/m at similar task success [20,23,24,30,35,37,42,43,48,57-61,63,65,67,70,73,74,76,78,80-86,89,93,94].

(3) Computation allocation & scheduling - 21/79 (26.6%):
Here, energy is reduced by placing work where/when it is cheaper: (a) offloading to peer robots or the cloud; (b) pipeline restructuring to shorten high-power bursts; (c) real-time scheduling to reduce idle/queueing; (d) batching and deadline-aware execution on embedded SoCs; (e) selecting accelerators (CPU/GPU/FPGA/TPU) under energy constraints. Studies typically report aggregate energy/power and runtime, sometimes with communication overhead accounted for [23,32,33,36,39,41,44,46,48,52,55,58,60,61,65,69,72,76,90,92,94].

(4) Power management & idle control - 9/79 (11.4%):

This family reduces non-productive consumption through (a) duty-cycling compute and sensors, (b) dynamic voltage/frequency scaling (DVFS), (c) sleep/standby during low-workload phases, (d) controller throttling, and (e) selective activation of perception/planning stacks. Outcomes are often expressed as average power reduction or longer runtime for a fixed task [29,32,33,52,58,62,77,79,95].

(5) Communication & data efficiency - 3/79 (3.8%):
Techniques cut radio energy via (a) selective transmission and compression/aggregation, (b) radio duty-cycling, and (c) positioning/link selection to increase channel gain and avoid costly re-transmissions. Studies usually trade small motion or compute costs for a net reduction in communication energy [26,27,62].

(6) Other - 1/79 (1.3%):
Idiosyncratic or cross-cutting approaches that do not cleanly map to the families above (e.g., one-off hardware-software co-tuning when the software method itself is not scheduling, power-management, motion, learning, or communication) [54].

Co-occurrence (qualitative): Multi-family designs are common: the most frequent pairings observed combine motion & trajectory with learning/prediction (e.g., learned energy-aware planners) and with computation scheduling (e.g., planners that co-design workload placement). Power-management was also seen paired with scheduling or communication to align sleep states with computational and radio activity.

Context vs. pre-2020: Swanborn & Malavolta's ten items map into these six families: advanced motion, limiting motion changes/speed → motion; limiting idle time → power management; offloading computation → computation allocation & scheduling; limiting unnecessary communication → communication & data efficiency; predicting task energy cost → learning/prediction; using optimized hardware → other when the contribution is primarily software-driven. The counts indicate that post-2020 work most often optimizes motion, with substantial use of learning/prediction and scheduling, which aligns with Section 3.3's finding that actuation is the dominant energy sink [2].

**Table 3: Technique Families Used to Improve Energy Efficiency**

| Family | k/79 (%) |
|---|---|
| Motion & trajectory optimization | 55 (69.6%) |
| Learning or predictive optimization | 32 (40.5%) |
| Computation allocation & scheduling | 21 (26.6%) |
| Power management & idle control | 9 (11.4%) |
| Communication & data efficiency | 3 (3.8%) |
| Other | 1 (1.3%) |

**Figure 7: Distribution of Techniques for Energy Efficiency; multi-label**

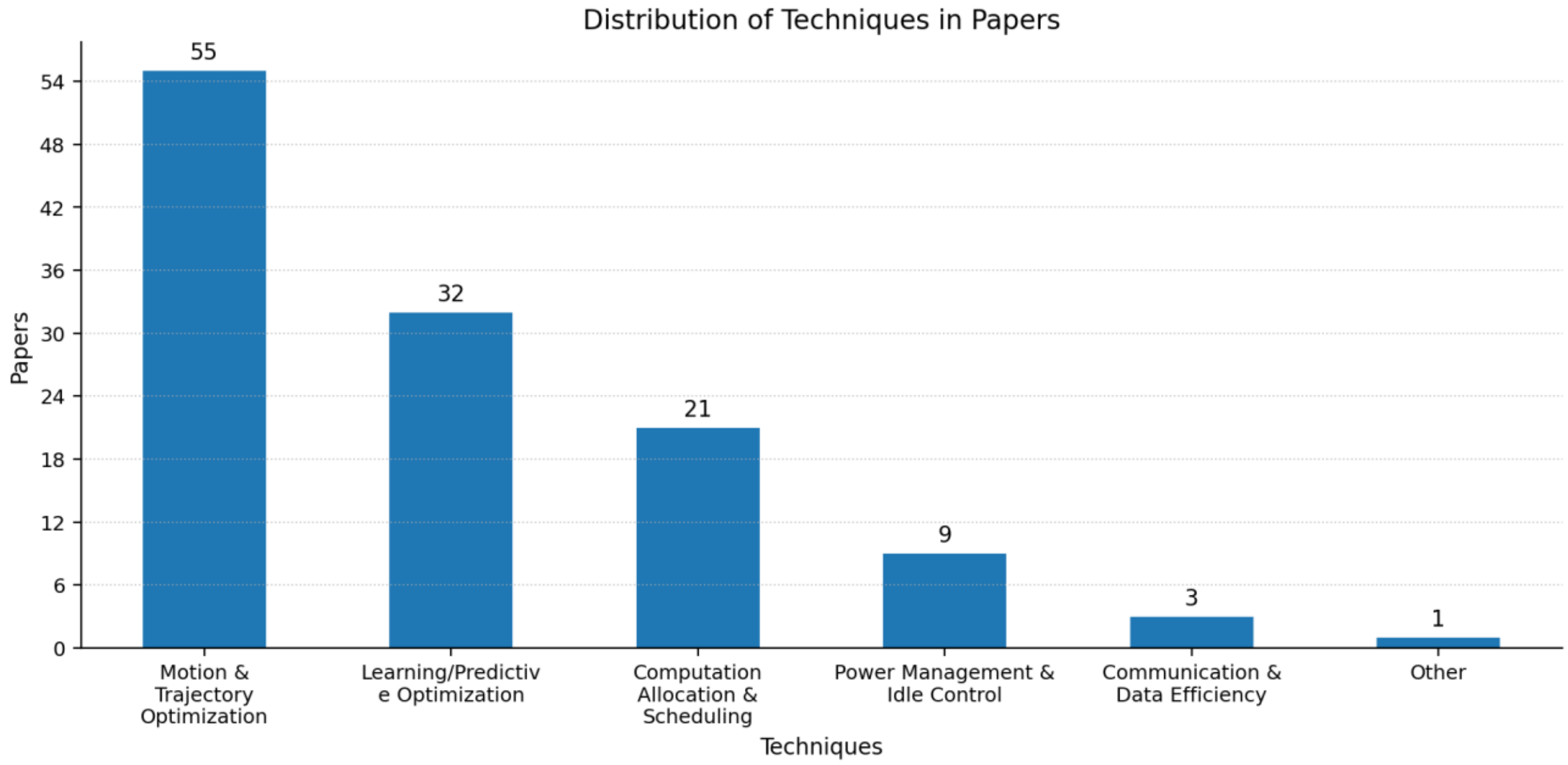

## 3.7 QA Trade-offs

For each study, the primary quality attribute (QA) that was analyzed in tension with energy was coded. The distribution is highly skewed toward performance vs. energy (74/79, 93.7%), with far fewer instances of accuracy vs. energy (3/79, 3.8%) and stability vs. energy (2/79, 2.5%).

Definitions used for coding:

- Performance vs. energy: time/throughput/path-length/latency contrasted with energy (e.g., saving energy by reducing speed, smoothing motion, or duty-cycling at the cost of longer completion time) [17-19,21-46,49,50,52-93,95].
- Accuracy vs. energy: task success or estimation accuracy (e.g., perception fidelity, localization error) measured against energy (e.g., fewer/cheaper sensor updates lower energy but may degrade accuracy) [20,51,94].
- Stability vs. energy: closed-loop stability or control quality (e.g., overshoot, settling time) contrasted with energy (e.g., aggressive energy saving causing oscillations or slower convergence) [47,48].

What drives the trade-offs:

- Techniques in motion & trajectory optimization (e.g., speed capping, fewer turns, smoother profiles) and power-management/idle control (e.g., sleep states, DVFS) most often instantiate the performance-energy tension: energy decreases, but execution time or latency typically rises [17-25,28,30,31,34-41,43-47,49-51,53,55,56,58-60,64-66,68,70-72,75,77,80,82-84,86-91,94,95].
- Computation allocation & scheduling sometimes introduces performance penalties (e.g., offloading adds transmission delay), yet can also improve both energy and time when it removes idle/queueing [23,32,33,36,39,41,44,46,48,52,55,58,60,61,65,69,72,76,90,92,94].
- Learning/predictive methods are mixed: some reject high-cost tasks (energy ↓, task throughput ↓), others learn policies that co-improve energy and performance by shortening paths or reducing replans [20,23,24,30,35,37,42,43,48,57-61,63,65,67,70,73,74,76,78,80-86,89,93,94].

Context vs. pre-2020: Swanborn & Malavolta also found performance to be the most common QA in tension with energy; the post-2020 corpus amplifies this pattern (74/79 vs. 8/17 pre-2020), while quantitative analyses of accuracy and stability remain infrequent [2].

**Figure 8: Distribution of QA Trade-offs**

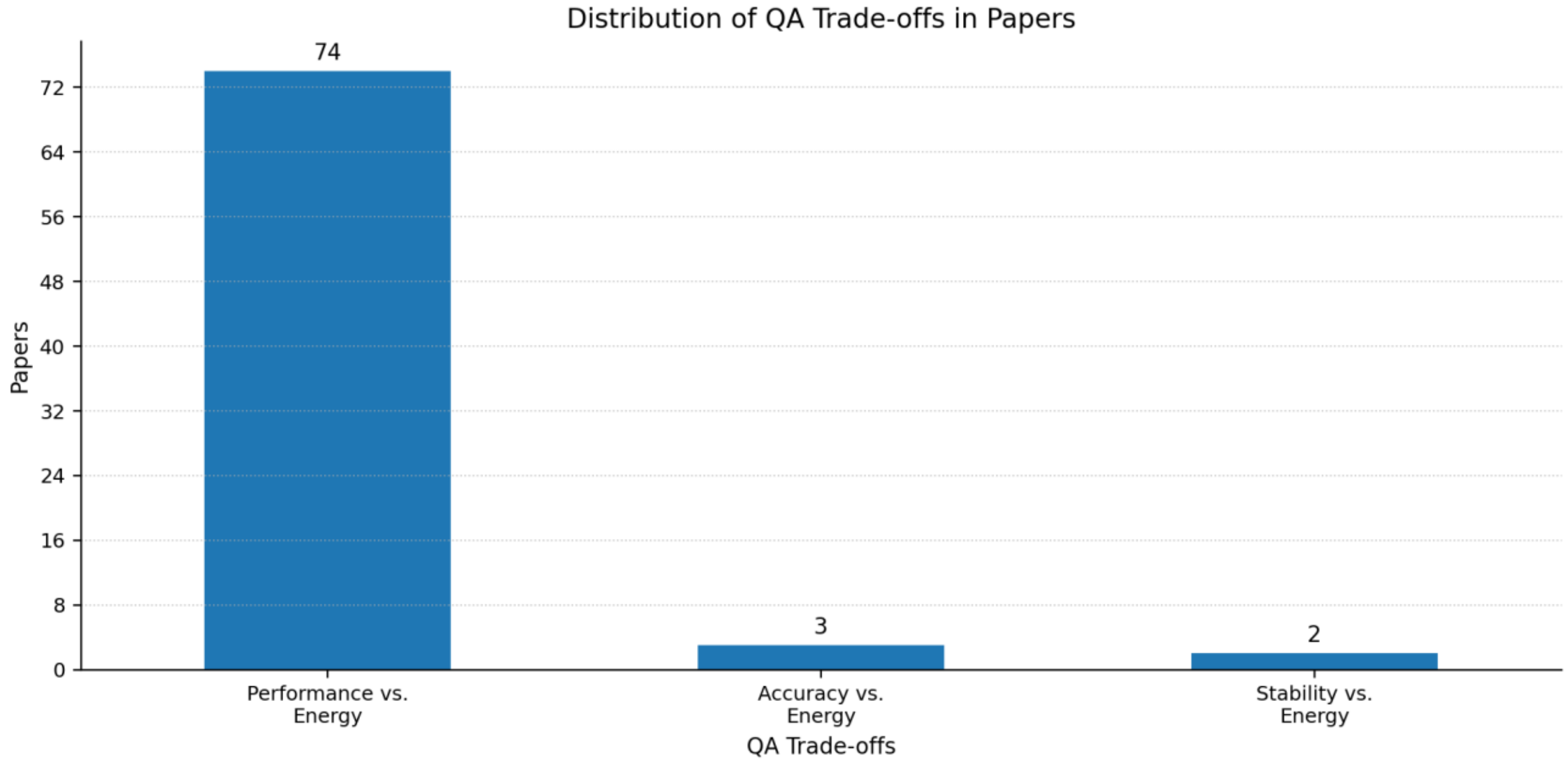

# 4. Discussion

## 4.1 Where the field stands post-2020

Three patterns dominate the literature. First, actuation remains the primary sink: motors/actuators are the main consumer in 68.4% of studies, with computing/controllers a distant second (13.9%). This aligns with the prevalence of motion & trajectory optimization (69.6%) as the leading technique family, often paired with learning/prediction (40.5%) and scheduling (26.6%).

Second, there is a clear tilt toward representational energy models (87.3%) and simulation-driven evaluation (51.9%). Compared with the ≤2020 corpus [2], this reflects stronger empirical grounding and model fidelity.

Third, metrics remain heterogeneous. Composite objectives that mix energy with other terms are most common (27 studies), followed by aggregate energy/power (21) and relative change vs. baseline (16); task-normalized measures (e.g., J/m, CoT) and performance-per-energy (e.g., FPS/W) are comparatively rare (4 each). This limits cross-paper comparability.

At the domain level, industrial applications now lead (31.6%), overtaking robot exploration (25.3%), which reverses the pre-2020 ranking [2]. This suggests growing attention to production contexts where even small percentage gains matter operationally.

Finally, the trade-off lens is overwhelmingly performance-centric: 93.7% of studies contrast energy with time/throughput/path length, while accuracy (3.8%) and stability (2.5%) are rarely quantified.

## 4.2 Implications for research and practice

Target actuation first, but mind the stack. Changes that reduce acceleration, steering effort, and stop-go behavior bring the largest savings; pairing them with lightweight perception/planning and sensible workload placement prevents compute/communication overheads from eroding gains.

Prefer representational models for design, report comparable metrics for evidence. Detailed models enable design-space exploration, but authors should also report task-normalized energy (e.g., J/m, J/(kg·m), CoT) alongside aggregate J/Wh or W to aid reuse in meta-analyses.

Plan for hybrid evaluation. A practical norm is emerging: prototype in simulation, then confirm at least one representative physical datapoint or dataset replay.

Make trade-offs explicit and proportional. Report Pareto-style points (energy vs. time/accuracy) instead of percentage savings alone; some reported savings come with large time penalties that are unacceptable for time-critical tasks.

## 4.3 Reporting practices to improve comparability

A minimal checklist is recommended for future work: (i) average power (W) and total energy (J/Wh) plus a task-normalized metric; (ii) clear baselines with absolute and relative results; (iii) explicit evaluation type (simulation/physical/hybrid); (iv) energy model category and key parameters (mass, friction, drivetrain losses, idle power); (v) multi-label technique families used; (vi) at least one quantified non-energy QA (time/throughput, accuracy, or stability).

## 4.4 Gaps and opportunities

Metrics standardization: Composite objectives dominate reporting; dual reporting (composite + comparable metric such as J/m or FPS/W) would lift comparability.

Under-explored families: Power management/idle control (11.4%) and communication/data efficiency (3.8%) are rare relative to their perceived importance in mobile robots; cross-layer designs that align motion plans with radio duty cycles and compute sleep windows are promising.

Beyond performance trade-offs: Only five studies quantify energy against accuracy or stability; energy-aware perception (sensor scheduling, resolution/frame-rate adaptation) and control-theoretic analyses of stability under DVFS/sleep policies remain open.

Industrial shift and benchmarks: With industrial now the top domain, benchmarks reflecting factory-like duty cycles (continuous operation, predictable layouts, safety constraints) would improve external validity of results beyond exploration settings.

From representational models to reproducibility: Sharing parameter sets and energy/power traces (e.g., current/voltage logs, payload, slip) would make results more reproducible and enable sensitivity analyses.

## 4.5 Limitations

The synthesis inherits limitations from the corpus and coding choices. Selection bias may favor positive or simulation-heavy results. Counting rules (e.g., one primary domain/consumer; multi-label techniques) shape distributions and may hide within-study nuance. Metric heterogeneity means some improvements are only comparable within a paper. Finally, “other” buckets in metrics and energy models aggregate disparate cases (including “no explicit model”), which may blur fine distinctions; captions and footnotes clarify scope.

(Context compared to ≤2020 relies on Swanborn & Malavolta's findings.)

## 5. Conclusions

This review synthesizes 79 post-2020 studies on energy-efficient robotics software. This study finds that (i) actuation dominates energy use, and most techniques therefore optimize motion, often coupled with learning and scheduling; (ii) representational energy models and empirical evaluations (including hybrid setups) are now the norm; (iii) metrics are diverse, with composite objectives common and task-normalized/performance-per-energy measures less frequent; and (iv) performance-energy is by far the most analyzed trade-off, while accuracy and stability are rarely quantified. These trends collectively indicate a maturing, empirically grounded field with clear opportunities for metric standardization, cross-layer optimization, and benchmarking in industrial settings.

Relative to the ≤2020 baseline [2], post-2020 work broadens the evidence base and strengthens evaluations, while reinforcing earlier insights about the primacy of actuation. Standardized reporting (Section 4.3) and broader analysis of non-performance trade-offs would further improve comparability and applicability.